\documentclass[sigconf,natbib]{acmart}
\usepackage{amsmath,amsfonts}
\usepackage{algorithmic}
\usepackage{booktabs}
\usepackage{balance}
\usepackage[english]{babel}
\usepackage{moresize}
\usepackage{amsmath}
\usepackage{algorithmic}
\usepackage[latin1]{inputenc}

\usepackage{comment}
\usepackage{paralist}
\usepackage{bm}
\usepackage{pgfplots}
\usetikzlibrary{pgfplots.dateplot}

\usepackage{flushend}
\usepackage[english]{babel}
\usepackage[latin1]{inputenc}
\usepackage{mathrsfs}
\usepackage{graphicx}

\usepackage{amssymb}
\usepackage{amsfonts}
\usepackage{url}
\usepackage{longtable}
\usepackage{rotating}
\usepackage{multirow}
\usepackage{mathrsfs}
\usepackage{subfigure}
\usepackage{enumitem}
\usepackage[linesnumbered,algoruled,boxed,lined]{algorithm2e}
\usepackage{adjustbox}
\usepackage{hyperref}
\usepackage{pgfplots}
\usetikzlibrary{pgfplots.dateplot}
\usepackage{filecontents}

\definecolor{tblue}{RGB}{31,119,180}
\definecolor{torange}{RGB}{255,127,14}
\definecolor{tgreen}{RGB}{44,160,44}
\definecolor{tred}{RGB}{214,39,40}
\definecolor{tpurple}{RGB}{148,103,189}

\newcommand{\hide}[1]{} 

\newcommand{\eg}{\textit{e}.\textit{g}.} 
\newcommand{\wrt}{\textit{w}.\textit{r}.\textit{t}}

\definecolor{tblue}{RGB}{31,119,180}
\definecolor{torange}{RGB}{255,127,14}
\definecolor{tgreen}{RGB}{44,160,44}
\definecolor{tred}{RGB}{214,39,40}
\definecolor{tpurple}{RGB}{148,103,189}

\def\model{HGAurban\xspace}

\begin{document}

\title{HGAurban: Heterogeneous Graph Autoencoding for Urban Spatial-Temporal Learning}

\author{Qianru Zhang}
\affiliation{
  \institution{The University of Hong Kong}
  \city{Hong Kong}
  \country{}
}
\email{zqrhku@connect.hku.hk}

\author{Xinyi Gao}
\affiliation{
  \institution{The University of Queensland}
  \city{Brisbane}
  \country{}
}

\author{Haixin Wang}
\affiliation{
  \institution{University of California, Los Angles}
  \city{Los Angles}
  \country{}
}

\author{Dong Huang}
\affiliation{
  \institution{National University of Singapore}
  \city{Singapore}
  \country{}
}

\email{}

\author{Siu-Ming Yiu}
\authornote{Corresponding author}
\affiliation{
  \institution{The University of Hong Kong}
  \city{Hong Kong}
  \country{}
}

\author{Hongzhi Yin}
\authornote{Corresponding author}
\affiliation{
  \institution{The University of Queensland}
  \city{Brisbane}
  \country{}
}

\begin{abstract}
Spatial-temporal graph representations play a crucial role in urban sensing applications, including traffic analysis, human mobility behavior modeling, and citywide crime prediction. 
However, a key challenge lies in the noisy and sparse nature of spatial-temporal data, which limits existing neural networks' ability to learn meaningful region representations in the spatial-temporal graph.
To overcome these limitations, we propose HGAurban, a novel heterogeneous spatial-temporal graph masked autoencoder that leverages generative self-supervised learning for robust urban data representation.
Our framework introduces a spatial-temporal heterogeneous graph encoder that extracts region-wise dependencies from multi-source data, enabling comprehensive modeling of diverse spatial relationships. Within our self-supervised learning paradigm, we implement a masked autoencoder that jointly processes node features and graph structure. This approach automatically learns heterogeneous spatial-temporal patterns across regions, significantly improving the representation of dynamic temporal correlations. 
Comprehensive experiments across multiple spatiotemporal mining tasks demonstrate that our framework outperforms state-of-the-art methods and robustly handles real-world urban data challenges, including noise and sparsity in both spatial and temporal dimensions.

\end{abstract}

\maketitle

\section{Introduction}
\label{sec:intro}

As remote sensing technologies and large-scale computing infrastructures continue to provide an abundance of spatial-temporal data, there is an increasing demand for robust spatial-temporal prediction frameworks in urban sensing applications. These applications include traffic analysis~\cite{STGODE,wang2020traffic,li2021spatial,wang2021secure,zhang2025autohformer,zhang2025fldmamba,wang2019origin,lv2018lc,zhang2025efficient}, as well as modeling human mobility behavior~\cite{pentland1999modeling,fuchs2023modeling, zhang2024asur,zhang2023online}. Additionally, citywide crime prediction~\cite{chen2021z,diao2019dynamic,long2023decentralized,zhang2025survey,zhang2024survey} is another important area of focus. A crucial aspect of developing effective spatial-temporal prediction frameworks lies in the accurate capture of spatial and temporal correlations between different geographical locations and time intervals. This ability ensures comprehensive insights into dynamic urban phenomena.

Researchers have introduced spatial-temporal graph neural networks (STGNNs) as a promising learning paradigm~\cite{sahili2023spatio,jin2023spatio,ta2022adaptive}. These networks are specifically designed to handle data with graph structures and have garnered significant attention due to their ability to capture multi-hop spatial and temporal dependencies. This is accomplished through the iterative propagation of messages among nodes, leveraging their knowledge of both spatial and temporal information. By employing this recursive message passing mechanism, STGNNs demonstrate exceptional proficiency in modeling intricate dependencies in spatial-temporal data. Their effectiveness lies in their capability to integrate and exploit the inherent structure and temporal dynamics of the data, enabling more accurate predictions and insightful analysis of various urban applications.

However, existing works in STGNNs have not adequately addressed the challenges of noise and sparsity in spatial-temporal data, which hinders the accurate modeling of dependencies. For instance, traffic data collected from sensors or GPS devices may contain outliers or missing values, leading to less relevant region connections and noisy signals. These noisy signals can negatively impact the performance of spatial-temporal prediction tasks, as they introduce irrelevant information and hinder the accurate modeling of dependencies. Besides, most existing solutions approach spatial-temporal prediction tasks in a supervised manner, relying on a sufficient amount of labeled data for training. However, spatial-temporal data is often sparsely labeled, especially for rare urban anomalies or events. This sparsity presents a significant challenge in building accurate prediction models. Thus, it becomes crucial to develop learning models that can effectively handle sparse labeled data and learn from limited samples for achieving accurate spatial-temporal predictions in real-world urban scenarios.

In light of these challenges, we propose \textbf{heterogeneous graph autoencoding for urban
spatial-temporal learning (\model)}. Our framework leverages multi-view spatial-temporal data and introduces a spatial-temporal heterogeneous graph neural encoder to capture region-wise dependencies. By considering both intra-view and inter-view data correlations, our model effectively models spatial-temporal patterns. Moreover, we propose a self-supervised learning paradigm that utilizes the mask mechanism to distill heterogeneous dependencies across observed regions in urban space. Specifically, we adopt the masked autoencoding mechanism on the node representations and graph structures to identify and emphasize the most relevant connections for prediction tasks. This enhances the learning process of implicit cross-region relations and improves the robustness of the model against noisy signals.

At the core of our approach is a generative spatial-temporal graph auto-encoder. This model learns to reconstruct both node features and the graph structure by utilizing the mask mechanism. By selectively masking certain regions in the input data during training, our model focuses on capturing the dependencies between regions and effectively reconstructing the missing or corrupted features and connections. This capability effectively overcomes the data sparsity by providing accurate self-supervisory signals to prioritize the most relevant information.

Our \model\ framework has been extensively evaluated in diverse spatial-temporal mining tasks and compared against multiple region representation and GNN baselines. The results demonstrate the effectiveness and superiority of our proposed method in addressing the challenges associated with data noise and sparsity. We highlight the key contributions as follows:
\begin{itemize}[leftmargin=*]

\item We place a strong emphasis on addressing the challenges of data noise in spatial-temporal prediction tasks. To achieve this, we introduce the superiority of self-supervised learning in region representation, incorporating the mask mechanism to enhance the robustness and generalization ability of our spatial-temporal heterogeneous graph neural architecture.

\item We propose a spatiotemporal heterogeneous graph neural network that captures both intra-view and inter-view regional dependencies while modeling diverse spatiotemporal factors. To address data sparsity, we introduce a masked autoencoding mechanism that masks both node features and graph structures within the heterogeneous spatiotemporal graph, enabling the model to effectively explore inter-correlations among regions.

\item We rigorously validate our \model\ across a range of spatial-temporal mining tasks, conducting comprehensive comparisons with state-of-the-art baselines. Our method consistently achieves the best performance across three diverse tasks. The code is publicly available at {\color{blue}{\href{https://github.com/lizzyhku/HGAurban__}{https://github.com/lizzyhku/\model}}}.
\end{itemize}

\section{Methodology}
\label{sec:solution}

\begin{figure*}
\centering
\includegraphics[width=1\linewidth]{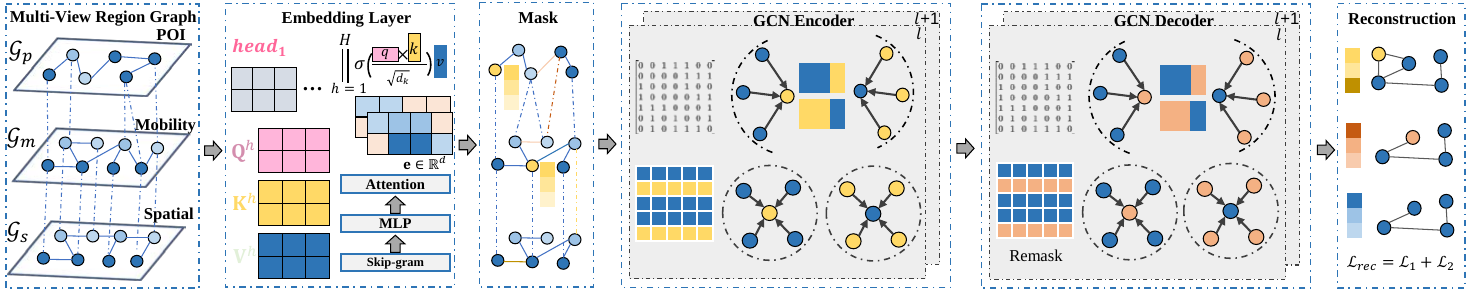}
\vspace{-0.2in}
\caption{Architecture of our \model\ region representation model with graph masked autoencoder.}
\label{fig:fra_01}
\end{figure*}

The framework of \model\ is depicted in Figure~\ref{fig:fra_01}, which addresses the task of reconstructing links and features of masked nodes using a mask mechanism based on generative self-supervised learning on the heterogeneous spatial-temporal graph. 
The \model\ framework operates on the heterogeneous spatial-temporal graph $\mathcal{G}$, which incorporates various types of information, such as spatial, temporal, and attribute data.
Our \model\ employs an encoder-decoder architecture. The encoder captures the underlying structure and derives meaningful representations from the heterogeneous spatiotemporal graph $\mathcal{G}$, while the decoder reconstructs the masked node features and graph links.
By utilizing the mask mechanism, \model\ learns to reconstruct the missing links and features of masked nodes, thereby capturing the correlations and dependencies. 
And the specific algorithm of \model\ is shown in Algorithm~\ref{alg:overall}.

\vspace{-0.1in}
\subsection{Preliminaries}
\label{sec:model}
In our methodology, we employ a partitioning approach to divide the geographic area of a city (such as New York) into $I$ spatial-regions. Each spatial-region is identified by an index, denoted as $i$ (e.g., $r_i$). To capture a wide range of urban contextual information, we propose a region embedding framework that integrates multiple diverse data sources associated with the metropolitan space. These data sources are incorporated into \model\ to generate enriched spatial-region representations, which are shown as follows.

\noindent \textbf{Region Point-of-Interests (POIs)}. We create a POI matrix $\mathbf{P}\in {\mathbb{R}^{I \times C}}$, where $C$ is the number of POI categories, to characterize the region's urban functions (such as entertainment and health).

\noindent \textbf{Human Mobility Trajectories}. We create a matrix $\mathbf{M}\in \mathbb{R}^{I \times T}$ utilizing  human trajectory to accurately represent the dynamic urban flows between citywide zones, where $T$ is the number of time slots. Individual human mobility traces are saved in this collection as $(r_s, r_d, t_s, t_d)$, where $r_s$ and $r_d$ denote source and destination regions, respectively, and $t_s$ and $t_d$ stand for timestamp data.

\noindent \textbf{Spatial Correlations}. We create a spatial matrix $\mathcal{D} \in \mathbb{R}^{I \times I}$. It is composed of region pairs like $(r_s, r_d, d')$ whose geographical distances are less than a threshold $\epsilon$, where $d'$ denotes the distance from the source node $r_s$ to the destination node $r_d$.

\noindent \textbf{Problem Statement:} Given the spatial-temporal graph, denoted as $\mathcal{G} = (\mathcal{V}, \mathcal{E})$, where $\mathcal{V}$ and $\mathcal{E}$ are node set and edge set respectively. $\mathcal{G}$ is built upon the POI matrix $\mathcal{P}$, the mobility matrix $\mathbf{M}$ and the spatial matrix $\mathcal{D}$. The task of this work is to learn a function $f$, which maps each node of $\mathcal{G}$ to the embedding space and reconstruct both the structure $\mathbf{A} \in \mathbb{R}^{|\mathcal{V}|\times |\mathcal{V}|}$ and features $\mathbf{X} \in \mathbb{R}^{|\mathcal{V}| \times D}$ of the input spatial-temporal graph $\mathcal{G}$. Meanwhile, the embedding space can effectively preserve the geographical and temporal connections between various locations and time slots, which is advantageous for a variety of urban sensing applications (\eg traffic forecasting and crime prediction).

\subsection{Embedding Layer}
To incorporate the Point-of-Interest (POI) context into region embeddings, we have developed a POI context embedding layer that reflects the functional information of each region in the latent representation space. Inspired by the POI token embedding technique using Skip-gram~\cite{rahmani2019category}, we input the region-specific POI vector into the Skip-gram model~\cite{cheng2006n} for POI context embedding. Subsequently, we concatenate the region-specific POI embeddings to generate POI-aware representations: $\textbf{E} = \text{MLP}(\text{Skip-gram}(\mathcal{P}))$.

Here, we employ a Multi-Layer Perceptron (MLP) for the embedding projection. The embedding table $\textbf{E}\in\mathbb{R}^{I\times d}$ represents the embeddings for all regions, where $d$ is the hidden dimensionality. Each region's embedding $\textbf{e}_i$ captures the region-specific POI contextual information in the latent semantic space. To incorporate region interactions within our POI embedding layer, we introduce a region-wise self-attention mechanism given by:
\begin{align}
    \label{eq:selfatt}
    \textbf{e}_i= \mathop{\Bigm|\Bigm|}_{h=1}^H \sum_{j=1}^I \alpha_{i,j}^h \cdot \textbf{V}^{h} \textbf{e}_j; ~~~
    \alpha_{i,j}^h = \delta\left(\frac{(\textbf{Q}^h\textbf{e}_i)^\top (\textbf{K}^h\textbf{e}_j)}{\sqrt{d/H}}\right)
\end{align}
Here, $\textbf{V}^h$, $\textbf{Q}^h$, and $\textbf{K}^h \in\mathbb{R}^{d/H\times d}$ correspond to the value, query, and key transformations for the $h$-th attention head, respectively. $H$ represents the number of attention heads, and $\mathop{\Bigm|\Bigm|}$ denotes the concatenation of the $H$ vectors. The softmax function $\delta(\cdot)$ is applied. In our self-attention layer, the embedding $\textbf{e}_i$ captures the pairwise interactions between region $r_i$ and $r_j$ with respect to the region-wise POI semantics.

\subsection{Heterogeneous Spatial-Temporal GNNs}
We model our data as a spatial-temporal graph $\mathcal{G}$, which is composed of three matrices, namely region POI matrix $\mathbf{P}$, human trajectory matrix $\mathbf{M}$ and spatial distance matrix $\mathcal{D}$. Thus the relations of the nodes in graph $\mathcal{G}$ are diverse and can be classified into five types, namely relations $\mathcal{R}_p$ based on POI similarities of regions, relations $\mathcal{R}_m$ based on urban flow between regions, relations $\mathcal{R}_d$ based on distances of regions, relations $\mathcal{R}_{pm}$ between the POI subgraph and the mobility subgraph, and relations $\mathcal{R}_{md}$ between the mobility subgraph and spatial distance graph. And the relation set $\mathcal{R}$ is shown as $\mathcal{R}=\{\mathcal{R}_p,\mathcal{R}_m, \mathcal{R}_d,\mathcal{R}_{pm},\mathcal{R}_{md}\}$. To handle the heterogeneous relationship, we design the relation-aware message passing paradigm as follows:
\begin{align}
    \textbf{h}_i^{(l+1)} = \sigma(\sum_{\gamma \in \mathcal{R}}\sum_{j\in\mathcal{N}^{\gamma}_i} \alpha_{i,\gamma} \textbf{W}_{\gamma}^{(l)} \textbf{h}_j^{(l)});~~
    \alpha_{i,\gamma} = \frac{1}{|\mathcal{N}^{\gamma}_i|}
\end{align}
\noindent where $\mathcal{N}^{\gamma}_i$ denotes neighbour set of region $r_i$ via relation $\gamma$. $\mathcal{R}$ is the relation set. $\textbf{h}_i^{(l+1)},\textbf{h}_j^{(l)}\in\mathbb{R}^d$ represents the hidden embedding vectors of the $i$-th region vertex in the $(l+1)$-th graph layer and the $j$-th region vertex in the $l$-th graph neural layer, respectively.  $\sigma(\cdot)$ denotes the ReLU activation function. $\alpha_{i,\gamma}\in\mathbb{R}$ represents the normalization weight of region vertex pair $(r_i, r_j)$, which is calculated using the degrees of the vertices.  
$\textbf{W}_{\gamma}^{(l)}\in\mathbb{R}^{d\times d}$ presents the learning weights for the $l-th$ layer. \model\ combines the multi-order embeddings to fully utilize the data obtained from the multi-hop neighborhood. The cross-view message passing mechanism is shown in the following formula:
\begin{align}
    \label{eq:multorder}
    \textbf{H} = \sum_{l=0}^L \textbf{H}^{(l)};~~~
    \textbf{H}^{(l+1)} = \sigma(\textbf{D}^{-\frac{1}{2}} \textbf{A} \textbf{D}^{-\frac{1}{2}} \textbf{H}^{(l)}\textbf{W}_{\gamma}^{(l+1)\top})
\end{align}
where $\textbf{H}\in\mathbb{R}^{|\mathcal{V}|\times d}$ is the embedding matrix, whose rows are regional embedding vectors $\textbf{h}_i$. $L$ is the number of the model layers. $\textbf{A}\in\mathbb{R}^{|\mathcal{V}|\times |\mathcal{V}|}$ denotes the adjacent matrix with self-loop. The diagonal degree matrix is denoted as $\textbf{D}\in\mathbb{R}^{|\mathcal{V}|\times |\mathcal{V}|}$.

\subsection{The Layout of Spatial-Temporal Graph Mask Autoencoder} 
Given the spatial-temporal graph $\mathcal{G}$, the encoder function $f_E$ and the decoder function $f_D$, Graph mask autoencoder (GMAE) aims to reconstruct the spatial-temporal graph $\mathcal{G}'$ as follows:
\begin{align}
\label{eq:gmae}
\textbf{H}  = f_E(\tilde{\mathbf{A}}, \tilde{\mathbf{X}}), ~~~~\mathcal{G}' = f_D(\tilde{\mathbf{A}}', \tilde{\mathbf{H}})
\end{align}
where $\tilde{\mathbf{A}}$ and $\tilde{\mathbf{X}}$ denote the masked adjacent matrix and mask node embedding matrix respectively. And $\tilde{\mathbf{A}'}$ and $\tilde{\mathbf{H}}$ represent the remasked adjacent matrix and the remasked node embedding matrix respectively. Details of GMAE design for the spatial-temporal graph $\mathcal{G}$ are in the following sections.

\subsubsection{Motivation of Features and Structure Reconstruction}
The primary motivation behind reconstructing features and structure in GAE lies in the versatility they offer to address different tasks and applications. In certain scenarios, an effective representation of the graph's structure $\mathbf{A}$ and node features $\mathbf{X}$ can then be utilized for tasks like link prediction or node classification. This can be formulated with coefficient $\alpha$ and activation $\sigma$ as follows:
\begin{equation}
\begin{matrix}
\underbrace{
\mathcal{L}
} \\ \text{Graph}
\end{matrix}
\quad
=
\quad
\begin{matrix}
\underbrace{
\Vert \mathbf{X} - \hat{\mathbf{X}} \Vert_F^2
} \\ \text{Node} \end{matrix}
+ \quad \alpha \cdot
\begin{matrix}
\underbrace{
\Vert \mathbf{A} - \sigma(\mathbf{H} \cdot \mathbf{H^T})} \Vert_F^2 \\ \text{Structure} \end{matrix}
\end{equation}
However, GAE often prioritizes reconstructing the features $\mathbf{X}$ while minimizing or ignoring the reconstruction error related to the structure $\mathbf{A}$. Actually, by adapting $\alpha$ based on the specific task and application, GAEs can effectively learn representations that cater to the demands of diverse graph-based problems.
Inspired by recent studies~\cite{park2019symmetric,salehi2019graph} and the significance of different types in our heterogeneous graph, we propose to reconstruct both the structure and features in our \model. 
Besides, we use a mask mechanism-based generative self-supervised learning paradigm, which allows for effective learning of the spatial-temporal dependencies of region-wise within the graph.

Jointly reconstructing both the structure and features offers several key advantages: 
\textbf{(1) Enhanced graph representation.} The model can achieve superior graph representation learning and node classification, which is particularly crucial for graphs with multiple relations.
\textbf{(2) Improved relationship nuance capture.} It can effectively captures the subtleties of the relationships between nodes, resulting in a more accurate representation of the graph structure. Such a detailed representation is challenging to achieve due to the complexity and multifaceted nature of node relationships.

\subsubsection{Reconstructing Features and Structures using Mask Mechanism}
Developing effective feature-oriented Graph Autoencoders (GAEs) is challenging due to the limited node feature dimensionality in graphs. Existing GAEs, primarily focused on feature reconstruction, often neglect the risk of learning the "identity function" when the hidden vector dimension surpasses the input, rendering the learned representation useless. To address this issue, we propose the utilization of denoising autoencoders, which intentionally introduce corruption to the input data. Masked autoencoders have demonstrated success in computer vision (CV) and natural language processing (NLP) tasks, where masking is employed as a corruption mechanism. In our suggested approach for GAEs, we advocate for the incorporation of masked autoencoders. Specifically, we randomly select a subset of nodes denoted as $\widetilde{\mathcal{V}} \in \mathcal{V}$ and mask their node features by utilizing a learned vector $x_M \in \mathbb{R}^d$. This strategy aims to prevent GAEs from learning the `identity function' and enhances the quality of the learned embeddings. By integrating masked autoencoders, we effectively address the challenge of feature reconstruction in GAEs and pave the way for more efficient and meaningful learning of feature-oriented graph representations. The masking process can be defined using Eq.~\ref{eq:mask_node}:
\begin{align}
x_i = \begin{cases}
x_M, & \text{if } v_i \in \widetilde{\mathcal{V}}\\
x_i, & \text{if } v_i \notin \widetilde{\mathcal{V}}
\end{cases}
\label{eq:mask_node}
\end{align}
Given the masked node feature matrix $\widetilde{\mathbf{X}}$, one of the objectives of \model\ is to reconstruct masked node features of regions (\eg, $r_i$) in $\widetilde{\mathcal{V}}$ and masked the adjacent matrix $\widetilde{\mathbf{A}}$ via a learned vector $a_M$.
In GNNs, the hidden embedding vector of each node is affected by its neighboring nodes' embedding vectors according to the message passing mechanism. However, selecting nodes via scores can lead to bias, for example, if all neighbors of one node are selected or masked. To prevent bias in node selection, random node selection or sampling is taken into account the entire graph structure rather than just local node properties, which also improves robustness.

\subsubsection{Remask Decoder via GNNs} The role of the decoder function $f_D$ is to map the hidden embedding vectors $\tilde{\mathbf{H}}$ to the input graph $\mathcal{G}'$ and its node embedding matrix $\mathbf{X}'$, based on the semantic level of the target input graph $\mathcal{G}$ and its corresponding node embedding matrix $\mathbf{X}$~\cite{he2022masked}. For example, in NLP, an MLP is typically sufficient for targets of one-hot missing words with rich semantics, as there is no structure semantics to consider. However, in CV, a transformer is often used as the decoder to recover pixel spots with few semantics.

To better capture graph structure information, we adopt GCN as the decoder in our method. This is in contrast to traditional GAEs that typically utilize an MLP to decode the hidden embedding matrix, which results in a similar output to the input graph embedding matrix. Using an MLP as the decoder cannot capture the structure semantics of graphs, so we choose GCN to better capture this.

To further improve the ability of the encoder to learn compressed graph representations, we adopt a remask mechanism in the decoder, inspired by~\cite{hou2022graphmae}. The process is represented as following:
\begin{align}\widetilde{h}_{i}=
\label{eq:remask}
\begin{cases}
h_{M}& ~~v_i \in \widetilde{\mathcal{V}}\\
h_i& v_i \notin \widetilde{\mathcal{V}}
\end{cases}\end{align}
Here, $\widetilde{\mathbf{H}} = REMASK(\mathbf{H})$, where a masked node is forced to build its input feature based on the neighboring non-masked nodes' hidden embedding vectors, under the guidance of the GCN decoder. This approach helps ensure that the learned embeddings are both informative and compressed, which can improve the overall region representation ability of our framework \model.

\subsection{Model Optimization and Complexity}
To optimize our method \model, we adopt the cosine similarity loss, denoted as $\mathcal{L}_1$, to optimize remasked node features and initial input node features. Besides, to help optimize structure masking, another loss MSE is adopted. These two losses are shown as: 
\begin{small}
 \begin{equation}
 \begin{aligned}
    \label{eq:loss1}
    \mathcal{L}_1 &= \frac{1}{N}\sum_{i=1}^N \left(  1 - \frac{h_i^T \, \widetilde{h}_{i}}{\Vert \widetilde{h}_{i} \Vert \cdot \Vert h_{i} \Vert} \right);\\
    \mathcal{L}_2 &= \left\Vert \mathbf{E} - \sigma \left( \widetilde{\mathbf{A}}' \cdot \widetilde{\mathbf{A}}'^T \right) \right\Vert_2^2
    \end{aligned}
\end{equation}
\end{small}
where $\mathbf{E}$ is the hidden matrix of initial adjacent matrix $\mathbf{A}$. 

By leveraging generative self-supervised learning tasks during pretraining, our proposed framework, \model, aims to produce high-quality region embeddings. The optimization goal of our framework \model\ can be formalized as follows:
\begin{align}
\label{eq:loss_all}
\min\limits_{\theta} \mathcal{L}_{rec}(\mathbf{A},\mathbf{X},\widetilde{\mathbf{A}}',\widetilde{\mathbf{X}}'),~~\mathcal{L}_{rec} = \mathcal{L}_1 + \mathcal{L}_2
\end{align}
where $\theta$ denotes the parameters of the model, $\mathcal{L}_{rec}$ is the reconstruction loss that measures the difference between the original and reconstructed adjacency matrix $\mathbf{A}$ and node feature matrix $\mathbf{X}$.
The objective function ensures that the model can accurately reconstruct both the graph structure and node features, while the second term encourages the model to learn representations that are useful for predicting node features in a self-supervised manner. Overall, the optimization goal of our framework reflects our aim to learn high-quality region embeddings that can be used for downstream tasks such as node classification, link prediction, and community detection. And the time complexity of our framework \model\ is divided into two parts: first of all, 
the multi-view space-time message passing paradigm has a complexity of $\mathcal{O}(|\mathcal{V}| \times L\times d)$, where $N$ and $L$ denote the number of edges and graph propagation layers, respectively. Secondly, for loss calculation, \model\ takes $\mathcal{O}(S\times |\tilde{\mathcal{V}}| \times d)$, where $S$ denotes the number of sampled nodes included in each batch for the spatial-temporal graph.

\subsection{In-depth Theory Analysis of \model}
In this section, we aim to provide in-depth analysis of our framework \model. We present a small \model\ loss denoting a small spatial-temporal alignment loss. Firstly, we have the following assumption: 
There exists a pseudo-inverse encoder $f_E$ which also the resulting pseudo autoencoder $s_g = \left\{g,f_g\right\}$satisfies $\mathbf{E}||s_g (z)-z|| \leq \epsilon$, where $z$ denotes masked representation of the input graph $\mathcal{G}$. Following this assumption, our method performs input-output alignment on the masked spatial-temporal graph. The lower bound of $\mathcal{L}_{rec} \geq \mathcal{L}_{asym}-\epsilon + C$, where $C$ is a constant number and $\mathcal{L}_{asym}$ is presented as following for the spatial-temporal graph $\mathcal{G}$:
\begin{align}
\label{eq:loss_asym}
\mathcal{L}_{asym} = -\mathbf{E}_{x_1,x_2}f(x_1)s_g(x_2) = -tr(\mathbf{H}_g^T \bar{\mathbf{A}}_m\mathbf{H})
\end{align}
where $\mathbf{H}$ is the output matrix of our method \model\ on $\mathcal{G}_1$ whose $x_1$-row is $\mathbf{H}_{x_1} = \sqrt{d_{x_1}}f(x_1)$, and $\mathbf{H}_g$ is the output matrix of $s_g$ on $\mathcal{G}_2$ whose $x_2$-th row is $(\mathbf{H}_g)_{x_2} = \sqrt{d_{x_2}}s_g(x_2)$. 
Thus, a small \model\ loss implies a small alignment loss as a lower bound, which also verifies align mask and unmask view via \model. This again verifies that our method can obtain the same level performance as the contrastive learning loss without data augmentation.

\begin{algorithm}[t]
\caption{\model: Heterogeneous Graph Autoencoding for Urban
Spatial-Temporal Learning}
\label{alg:overall}
\begin{algorithmic}[1]
\REQUIRE 
    \STATE Spatio-temporal graph $\mathcal{G}=(\mathcal{V},\mathcal{E})$ with:
    \STATE \quad POI matrix $\mathcal{P} \in \mathbb{R}^{I \times C}$
    \STATE \quad Mobility matrix $\mathbf{M} \in \mathbb{R}^{I \times I \times T}$
    \STATE \quad Distance matrix $\mathcal{D} \in \mathbb{R}^{I \times I}$
    \STATE Mask ratio $p_{\text{mask}} \in (0,1)$
    \STATE Relation set $\mathcal{R}=\{\mathcal{R}_p,\mathcal{R}_m,\mathcal{R}_d,\mathcal{R}_{pm},\mathcal{R}_{md}\}$
    
\ENSURE Reconstructed adjacency $\hat{\mathbf{A}}$ and features $\hat{\mathbf{X}}$

\STATE \textbf{/* Graph Partitioning */}
\STATE Divide urban area into $I$ regions $\{r_i\}_{i=1}^I$
\STATE Initialize node features $\mathbf{X} \leftarrow \text{Concat}(\mathbf{P}, \mathbf{M}, \mathcal{D})$

\STATE \textbf{/* Masked Autoencoder Setup */}
\STATE Randomly select $\mathcal{V}_{\text{mask}} \subset \mathcal{V}$ with $|\mathcal{V}_{\text{mask}}| = p_{\text{mask}}|\mathcal{V}|$
\STATE Generate masked graph $\tilde{\mathcal{G}}$ via Eq.~\ref{eq:mask_node} and Eq.~\ref{eq:remask}

\STATE \textbf{/* Encoder: Heterogeneous GNN */}
\FOR{$l=1$ \TO $L$}
    \STATE Multi-relational message passing:
    \STATE $\mathbf{h}_i^{(l+1)} \leftarrow \sigma\Big(\sum_{\gamma \in \mathcal{R}}\sum_{j\in\mathcal{N}^{\gamma}_i} \frac{1}{|\mathcal{N}^{\gamma}_i|} \mathbf{W}_{\gamma}^{(l)} \mathbf{h}_j^{(l)}\Big)$
\ENDFOR
\STATE $\mathbf{H} \leftarrow \sum_{l=0}^L \mathbf{H}^{(l)}$ \COMMENT{Final embeddings}

\STATE \textbf{/* Decoder: Reconstruction */}
\STATE Remask embeddings $\tilde{\mathbf{H}} \leftarrow \text{REMASK}(\mathbf{H})$ (Eq.~\ref{eq:remask})
\STATE $\hat{\mathbf{X}} \leftarrow \text{GCN-Decoder}(\tilde{\mathbf{H}})$
\STATE $\hat{\mathbf{A}} \leftarrow \sigma(\mathbf{H}\mathbf{H}^\top)$

\STATE \textbf{/* Loss Computation */}
\STATE Feature loss: $\mathcal{L}_1 \leftarrow \frac{1}{N}\sum_{i=1}^N \left(1 - \cos(\mathbf{h}_i, \tilde{\mathbf{h}}_i)\right)$
\STATE Structure loss: $\mathcal{L}_2 \leftarrow \|\mathbf{A} - \sigma(\tilde{\mathbf{A}}' \tilde{\mathbf{A}}'^\top)\|_F^2$
\STATE Total loss: $\mathcal{L} \leftarrow \mathcal{L}_1 + \mathcal{L}_2$

\RETURN $\hat{\mathbf{A}}, \hat{\mathbf{X}}$
\end{algorithmic}
\end{algorithm}

\begin{table}[h]
\centering
\footnotesize
\setlength{\tabcolsep}{0.36mm}{
\caption{Data Description of Experimented Datasets}
\vspace{-2mm}
\label{fig:data_sta}
\begin{tabular}{|l|c|c|}
\hline
\textbf{Data}        & \textbf{Description of Chicago Data}                                                                                        & \textbf{Description of NYC data} \\ 
\hline 
Census  &\begin{tabular}[c]{@{}c@{}}Boundaries of 234 regions \\split by streets in \\a certain district, Chicago\end{tabular}         &\begin{tabular}[c]{@{}c@{}}Boundaries of 180 regions\\ split by streets in\\ Manhattan, New York\end{tabular}                             \\ \hline 
Taxi trips & \begin{tabular}[c]{@{}c@{}}Total 386,272 taxi \\trips during a month\end{tabular}                  &\begin{tabular}[c]{@{}c@{}} Total 1,445,285 taxi \\trips during a month\end{tabular}                             \\ \hline
Crime data     &\begin{tabular}[c]{@{}c@{}}Total 321,876 \\crime records during 1 year\end{tabular}   &\begin{tabular}[c]{@{}c@{}}Total 108,575 crime\\ records during 1 year\end{tabular}                             \\ \hline
POI data       &\begin{tabular}[c]{@{}c@{}} Total 3,680,125 POI locations\\ of 130 categories \end{tabular}  &\begin{tabular}[c]{@{}c@{}}Total 20,569 POI locations\\ of 50 categories\end{tabular} 
\\ \hline 
House price       &\begin{tabular}[c]{@{}c@{}}Total 44,447 house price data\\
in a certain district, Chicago\end{tabular}  &\begin{tabular}[c]{@{}c@{}}Total 22,540 house price data\\
in Manhattan, New York\end{tabular} 
\\  \hline
\end{tabular}}
\vspace{-0.5cm}
\end{table}

\vspace{-0.1cm}
\section{Experiments}
\label{sec:eval}
We conduct experiments to answer the following questions: \textbf{RQ1:} How does the \model\ fare in terms of different baselines across multiple spatial-temporal learning applications including traffic prediction, crime prediction and house price prediction?
\textbf{RQ2:} How do various data perspectives and generative elements impact the effectiveness of region representation?
\textbf{RQ3:} How well does our \model\ fare in representation learning over areas with various levels of data sparsity?
\textbf{RQ4:} What effects do different hyperparameter settings have on the performance of the \model's region representation?
\textbf{RQ5:} How the efficiency our method \model\ than other region representation methods?

\vspace{-0.1in}
\subsection{Experimental Setup}
\label{sec:setup}

\subsubsection{\bf Datasets and Protocols} With the aim of assessing the effectiveness of our \model\ framework, we gathered real-world datasets from two major cities, namely Chicago and New York City. These datasets were utilized to perform three spatio-temporal mining tasks: crime prediction and traffic flow forecasting, as well as property price prediction. To ensure a comprehensive evaluation, we included various types of crimes in our datasets. For the Chicago database, crimes such as Theft, Battery, Assault, and Damage were considered, while the NYC database encompassed crimes such as Burglary, Larceny, Robbery, and Assault. The selection of crime types aligns with the parameters in \cite{xia2021spatial}. A detailed description of the data is shown in Table~\ref{fig:data_sta}.

\subsubsection{\bf Hyperparameter Settings}
Following existing region representation studies~\cite{wu2022multi_graph,zhang2021multi}, we set the dimensionality $d$ as 96. And from the hyperparameter study experiments, we find that the framework \model\ obtains the best performance when the GCN depth is set as 2. And the learning rate is 0.001 with weight decay 0.0005. For crime prediction backbone model, the learning rate of ST-SHN
is set as 0.001 and weight decay is 0.96. The depth of the spatial aggregation layer is set as 2. For traffic prediction backbone model ST-GCN, the input length is 12 time intervals with each interval setting as 5 minutes. And the output is 15 minutes' results of traffic flow prediction.

\subsubsection{\bf Implementation Details}
To be fair, we implement \model\ and other baselines in Python 3.8, pytorch 1.7.0 with cuda 11.3 for our method implementation and tensorflow 1.15.3 for traffic prediction and crime prediction. The experiments are conducted on a server with 10-cores of Intel(R) Core(TM) i9-9820X CPU @ 3.30GHz 64.0GB RAM and one Nvidia GeForce RTX 3090 GPU.

\begin{table*}[t]
\center
\setlength{\abovecaptionskip}{0cm}
\setlength{\belowcaptionskip}{0cm}
\setlength{\tabcolsep}{4.5pt}
\footnotesize
\vspace{0.3cm}
\caption{Overall performance in crime forecasting, traffic prediction, and house price prediction.}
\label{table:overall}
\resizebox{\textwidth}{!}{
\begin{tabular}{|l|c|c|c|c|c|c|c|c|c|c|c|c|c|c|}
    \hline
    \multirow{3}{*}{Model} & \multicolumn{4}{c|}{Crime Prediction} & \multicolumn{6}{c|}{Traffic Prediction} & \multicolumn{4}{c|}{House Price Prediction} \\
    \cline{2-15}
    & \multicolumn{2}{c|}{CHI-Crime} & \multicolumn{2}{c|}{NYC-Crime} & \multicolumn{2}{c|}{CHI-Taxi} & \multicolumn{2}{c|}{NYC-Bike} & \multicolumn{2}{c|}{NYC-Taxi} & \multicolumn{2}{c|}{CHI-House} & \multicolumn{2}{c|}{NYC-House}\\ 
    \cline{2-15}
    & MAE $\downarrow$ & MAPE $\downarrow$ & MAE $\downarrow$ & MAPE $\downarrow$ & MAE $\downarrow$ & RMSE $\downarrow$ & MAE $\downarrow$ & RMSE $\downarrow$ & MAE $\downarrow$ & RMSE $\downarrow$ & MAE $\downarrow$ & MAPE $\downarrow$ & MAE $\downarrow$ & MAPE $\downarrow$\\ \cline{2-15} 
    \hline \hline
    ST-SHN &2.0259 &0.9987 &4.4004 &0.9861 & -- & -- & -- & -- & -- & -- & -- & -- & -- & --  \\
    \hline
    ST-GCN & -- & -- & -- & -- & \multicolumn{1}{c|}{0.1395} &0.5933 & \multicolumn{1}{c|}{0.9240} & 1.8562 & \multicolumn{1}{c|}{1.4093} & 4.1766 & -- & -- & -- & -- \\
    \hline
    Node2vec &1.6334  &0.8605  &4.3646  &0.9454  & \multicolumn{1}{c|}{0.1206} & 0.5803 & \multicolumn{1}{c|}{0.9093} & 1.8513 & \multicolumn{1}{c|}{1.3508} & 4.0105 & \multicolumn{1}{c|}{13137.2178}    &\multicolumn{1}{c|}{44.4278}  & \multicolumn{1}{c|}{4832.6905}    &\multicolumn{1}{c|}{19.8942} \\ 
    \hline
    GCN &1.6061  &0.8546  &4.3257  &0.9234  & \multicolumn{1}{c|}{0.1174} & 0.5707 & \multicolumn{1}{c|}{0.9144} & 1.8321 & \multicolumn{1}{c|}{1.3819} & 4.0200 & \multicolumn{1}{c|}{13074.2121} & \multicolumn{1}{c|}{42.6572} & \multicolumn{1}{c|}{4840.7394} & \multicolumn{1}{c|}{18.3315}\\
    \hline
    GAT &1.5742 &0.8830 &4.3455 &0.9267 & \multicolumn{1}{c|}{0.1105} & 0.5712 & \multicolumn{1}{c|}{0.9110} & 1.8466 & \multicolumn{1}{c|}{1.3746} & 4.0153 & \multicolumn{1}{c|}{13024.7843} &\multicolumn{1}{c|}{43.3221} & \multicolumn{1}{c|}{4799.8482} & 18.3433 \\ 
    \hline
    GraphSage &1.5960 &0.8713 &4.3080 &0.9255 & \multicolumn{1}{c|}{0.1196} & 0.5796 & \multicolumn{1}{c|}{0.9102} & 1.8473 & \multicolumn{1}{c|}{1.3966} & 4.0801 & \multicolumn{1}{c|}{13145.5623} & \multicolumn{1}{c|}{44.3167} & \multicolumn{1}{c|}{4875.6026} & \multicolumn{1}{c|}{18.4570} \\ 
    \hline 
    GAE &1.5711 &0.8801 &4.3749 &0.9343 & \multicolumn{1}{c|}{0.1103} & 0.5701 & \multicolumn{1}{c|}{0.9132} & 1.8412 & \multicolumn{1}{c|}{1.3719} & 4.0337 & \multicolumn{1}{c|}{13278.3256} & \multicolumn{1}{c|}{42.3221} & \multicolumn{1}{c|}{4896.9564} & \multicolumn{1}{c|}{18.3114}\\
    \hline
     GraphCL &1.2332 &0.6293 &3.3075 &0.6771 & \multicolumn{1}{c|}{0.0812} &0.5364  & \multicolumn{1}{c|}{0.8582} &1.8180  & \multicolumn{1}{c|}{1.3022} &3.7029  & \multicolumn{1}{c|}{10752.5693} & \multicolumn{1}{c|}{28.8374} & \multicolumn{1}{c|}{4562.7279} & \multicolumn{1}{c|}{11.3055}\\
    \hline
    RGCL &1.1946 &0.6081 &3.1026 &0.6323 & \multicolumn{1}{c|}{0.0797} &0.5052  & \multicolumn{1}{c|}{0.8319} &1.8107  & \multicolumn{1}{c|}{1.2973} &3.6865  & \multicolumn{1}{c|}{10673.3289} & \multicolumn{1}{c|}{27.5279} & \multicolumn{1}{c|}{4439.0733} & \multicolumn{1}{c|}{10.2375}\\
    \hline
    POI &1.3047 &0.8142 &4.0069 &0.8658 & \multicolumn{1}{c|}{0.0933} & 0.5578 & \multicolumn{1}{c|}{0.8892} & 1.8277 & \multicolumn{1}{c|}{1.3316} & 3.9872 & \multicolumn{1}{c|}{12045.3212} & \multicolumn{1}{c|}{33.5049} & \multicolumn{1}{c|}{4703.3755} & \multicolumn{1}{c|}{16.7920}\\ 
    \hline
    HDGE &1.3586 &0.8273 &4.2021 &0.7821 & \multicolumn{1}{c|}{0.0865} & 0.5502 & \multicolumn{1}{c|}{0.8667} & 1.8251 & \multicolumn{1}{c|}{1.2997} & 3.9846 & \multicolumn{1}{c|}{11976.3215} & \multicolumn{1}{c|}{30.8451} & \multicolumn{1}{c|}{4677.6905} & \multicolumn{1}{c|}{12.5192}\\ 
    \hline
    ZE-Mob &1.3954 &0.8249 &4.3560 &0.8012 & \multicolumn{1}{c|}{0.1002} & 0.5668 & \multicolumn{1}{c|}{0.8900} & 1.8359 & \multicolumn{1}{c|}{1.3314} & 4.0366 & \multicolumn{1}{c|}{12351.1321} & \multicolumn{1}{c|}{38.6171} & \multicolumn{1}{c|}{4730.6927} & \multicolumn{1}{c|}{16.2586}\\ 
    \hline
    MV-PN &1.3370 &0.8132 &4.2342 &0.7791 & \multicolumn{1}{c|}{0.0903} & 0.5502 & \multicolumn{1}{c|}{0.8886} & 1.8313 & \multicolumn{1}{c|}{1.3306} & 3.9530 & \multicolumn{1}{c|}{12565.0607} & \multicolumn{1}{c|}{39.7812} & \multicolumn{1}{c|}{4798.2951} & \multicolumn{1}{c|}{17.0418}\\ 
    \hline
    CGAL &1.3386 &0.7950 &4.1782 &0.7506 & \multicolumn{1}{c|}{0.1013} & 0.5682 & \multicolumn{1}{c|}{0.9097} & 1.8557 & \multicolumn{1}{c|}{1.3353} & 4.0671 & \multicolumn{1}{c|}{12094.5869}    &\multicolumn{1}{c|}{36.9078}     & \multicolumn{1}{c|}{4731.8159}    &\multicolumn{1}{c|}{16.5454}\\
    \hline
    MVURE &1.2586 &0.7087 &3.7683 &0.7318 & \multicolumn{1}{c|}{0.0874} & 0.5405 & \multicolumn{1}{c|}{0.8699} & 1.8157 & \multicolumn{1}{c|}{1.3007} & 3.6715 & \multicolumn{1}{c|}{11095.5323}    &\multicolumn{1}{c|}{34.8954}     & \multicolumn{1}{c|}{4675.1626}    &\multicolumn{1}{c|}{15.9860}\\ 
    \hline
    AutoST & 1.0480 & 0.4787 & 2.1073 & 0.5241 & 0.0665 & 0.4931 & 0.8364 & 1.8145 & 1.2871 & 3.6446 & 10463.7715 & 25.7575 & 4517.7276 & 7.1660\\ \hline
    MGFN &1.2538 &0.6937 &3.5971 &0.7065 & \multicolumn{1}{c|}{0.0831} & 0.5385 & \multicolumn{1}{c|}{0.8783} & 1.8163 & \multicolumn{1}{c|}{1.3266} & 3.7514 & \multicolumn{1}{c|}{10792.7834}    &\multicolumn{1}{c|}{29.9832}    & \multicolumn{1}{c|}{4651.3451}    &\multicolumn{1}{c|}{12.9752}\\\hline\hline
    \textbf{\model} & \textbf{1.0438} & \textbf{0.4608} & \textbf{2.1060} & \textbf{0.5203} & \textbf{0.0606} & \textbf{0.4817} & \textbf{0.7787} & \textbf{1.8079} & \textbf{1.2778} & \textbf{3.6335} & \textbf{10375.9982} & \textbf{23.1751} & \textbf{4432.1934} & \textbf{6.8785}\\
    \hline
    P-Value & 5.27e-12 & 1.74e-14 & 2.7e-13 & 8.22e-11 & 1.32e-6 & 7.34e-10 & 2.13e-10 & 9.38e-4 & 1.24e-18 & 5.37e-21 & 2.18e-5 & 1.78e-4 &1.91e-4 & 2.75e-4\\
    \hline
\end{tabular}}
\label{tab:over_results}
\vspace{-0.2in}
\end{table*}

\subsubsection{\bf Baselines}
\label{app:baseline}
We use 15 baselines from three research lines namely, graph representation approaches, graph contrastive learning models, and spatial-temporal region representation methods for comparison in order to thoroughly assess our \model\ model.\\
\noindent \textbf{Graph Representation Approaches}:
\begin{itemize}[leftmargin=*]
\item \textbf{Node2vec}~\cite{grover2016node2vec} encodes graph structural data using a random walk-based skip-gram. 
\item \textbf{GCN}~\cite{kipf2016semi} carries out the convolution-based message transfer along the edges between neighboring nodes for refinement. 
\item \textbf{GraphSage}~\cite{hamilton2017inductive} is a graph neural architecture that permits the accumulation of information from the sub-graph structures.
\item \textbf{GAE}~\cite{kipf2016variational} maps node into latent embedding space by the Graph Auto-encoder using the input reconstruction objective. 
\item  \textbf{GAT}~\cite{velivckovic2017graph} varies the degrees of significance among nearby nodes, and improves GNNs' capacity to discriminate between diverse inputs. To optimize the performance of GNN-based approaches, the number of information propagation layers is adjusted from a range of \{1,2,3,4\}.
\end{itemize}

\noindent \textbf{Graph Contrastive Learning Models}:
\begin{itemize}[leftmargin=*]
\item \textbf{GraphCL}~\cite{you2020graph} is based on the maximizing of mutual knowledge and generates many contrastive viewpoints for augmentation.
\item \textbf{RGCL}~\cite{li2022let} have been fine-tuned to manage the contrastive augmentation in order to allow for fair comparison.
\end{itemize}

\noindent \textbf{Spatial-Temporal Region Representation Methods}:
\begin{itemize}[leftmargin=*]
\item \textbf{POI}~\cite{rahmani2019category} uses TF-IDF tokenization with the provided POI matrix and the POI properties to represent spatial regions.

\item \textbf{HDGE}~\cite{wang2017region} creates a crowd flow graph using human trajectory data and embeds regions into latent vectors to protect the structural data of the network and capture mobility patterns. 

\item \textbf{ZE-Mob}~\cite{yao2018representing} uses region correlations to create embeddings while taking into account human movement and taxi moving traces based on multi-view data sources. 
\item \textbf{MV-PN}~\cite{fu2019efficient} models intra-regional and inter-regional correlations with an encoder-decoder network.
\item \textbf{CGAL}~\cite{zhang2019unifying} is a technique for graph-regularized adversarial learning that uses pairwise graph relationships to embed regions in latent space via the graph attention neural network.
\item \textbf{MVURE}~\cite{zhang2021multi} makes use of the graph attention mechanism to simulate region correlations with the built-in properties of the region and data on human mobility.
\item \textbf{AutoST}~\cite{zhang2023automated} is a method that utilizes contrastive self-supervised learning method via multi-view graph to model region representation learning problem on traffic and crime prediction tasks.  
\item \textbf{MGFN}~\cite{wu2022multi_graph} aggregates information for both intra-pattern and inter-pattern patterns, and encodes region embeddings with multi-level cross-attention without region function information.
\end{itemize}
 \vspace{-0.1in}

\begin{figure*}[t]
\centering
\begin{tabular}{c c c c }
\hspace{40.5mm}
\begin{minipage}{0.5cm}
\includegraphics[width=7.5cm]{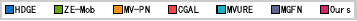}
\end{minipage}\vspace{0.5mm}\hspace{-10.5mm}
&
\\\hspace{-4.0mm}
  \begin{minipage}{0.2\textwidth}
	\includegraphics[width=\textwidth]{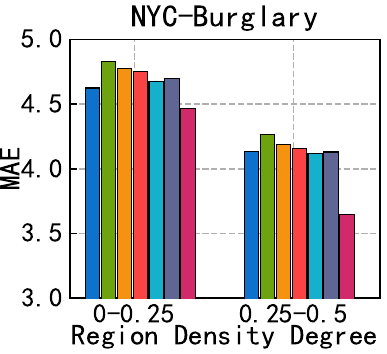}
  \end{minipage}\hspace{-3.5mm}
  &
  \begin{minipage}{0.2\textwidth}
	\includegraphics[width=\textwidth]{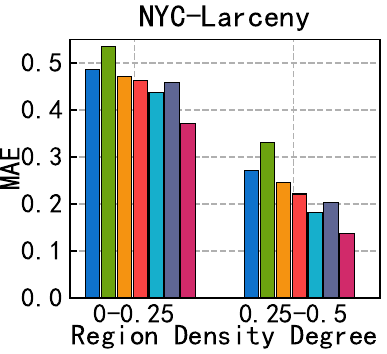}
  \end{minipage}\hspace{-3.5mm}
  &
    \begin{minipage}{0.2\textwidth}
	\includegraphics[width=\textwidth]{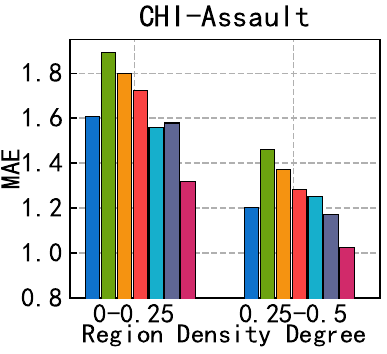}
  \end{minipage}\hspace{-3.5mm}

  &
  \begin{minipage}{0.2\textwidth}
	\includegraphics[width=\textwidth]{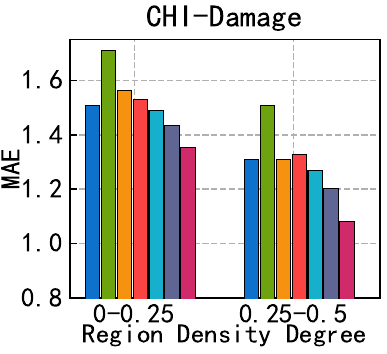}
  \end{minipage}\hspace{-3.5mm}
\end{tabular}
\caption{Results on NYC and CHI crime for four crime types \wrt\ different data density degrees.}
\label{fig:robustness1}
\end{figure*}

\subsection{Effectiveness Evaluation (RQ1)}
Table~\ref{table:overall} shows a performance comparison of all methods on three tasks (crime prediction, traffic prediction, and house price prediction) in terms of MAE, MAPE, and RMSE.

\noindent\textbf{Crime prediction}. We show crime prediction results in Table~\ref{table:overall}. From Table~\ref{table:overall}, we summarize observations as follows: \vspace{-0.05in}
\begin{itemize}[leftmargin=*]
\item On both datasets, our framework \model\ performs best at predicting crimes, indicating the viability of our generative spatio-temporal graph self-supervised learning approach. In specific, we attribute the benefits to: i) \model\ can capture both the mutli-view area relationships from multi-typed data sources (i.e., POI semantics, human motion traces, geographical coordinates) thanks to masking different types of relations. ii) The adaptive data augmentation way of \model\ via the mask mechanism on nodes and relations improves the representation ability against noise disturbance and skewed data distribution. 
\item While existing approaches for region representation (such as CGAL, MVURE, and MGFN) aim to capture the mobility-based region correlations, they are susceptible to noisy edges in their built mobility-based region graphs. The performance of region representation learning will be hampered by the propagation of information propagation between less-relevant regions. Besides, the spatial dependence modeling in such baselines is constrained by the skewed distribution of mobility data. Our \model\, which is complementary to the mobility-based region connections, adds region self-discrimination supervision signals to the region embedding paradigm to solve these limitations.
\item The large enhancement of \model\ over the compared graph embedding techniques (\textit{i.e.,} GCN and GAT) further supports the significance of generative self-supervised learning paradigm. When a message passes over the edges of a task-irrelevant region in the graph neural paradigm, noise effects will be enhanced. The information compiled from all the nearby mobility-based or POI-based graph regions may be misleading when it comes to encoding the actual underlying crime trends.
\end{itemize}

\noindent\textbf{Traffic prediction}. We present traffic prediction results in Table~\ref{table:overall} with the following observations: 
\begin{itemize}[leftmargin=*]
\item The significance of the generative self-supervised learning with the mask mechanism is further supported by the significant performance improvement of \model\ over the compared graph embedding techniques (such as GCN, GAT, and GAE). In the graph neural paradigm, noise effects will be amplified when messages crosses edges of regions that are unrelated to the task. 
\item To find traffic patterns or mine human behavior, region representation methods (\textit{i.e.,} MVURE, MGFN) need enough data, which means they cannot handle long-tail data. Contrarily, contrastive learning with masking nodes or links mechanism has better skewed data distribution adaption. 
\end{itemize}

\begin{figure*}[t]
\centering
\begin{tabular}{c c c c}
\hspace{-6mm}
  \begin{minipage}{0.24\textwidth}
	\includegraphics[width=\textwidth]{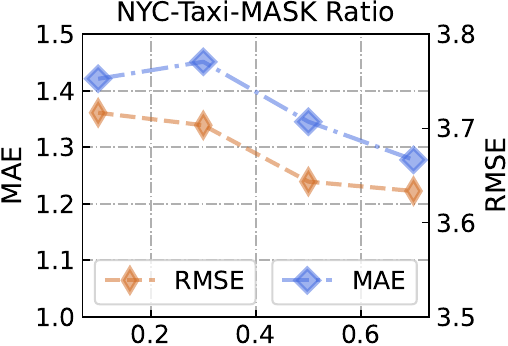}
  \end{minipage}\hspace{-3.0mm}
 &
  \begin{minipage}{0.24\textwidth}
    \includegraphics[width=\textwidth]{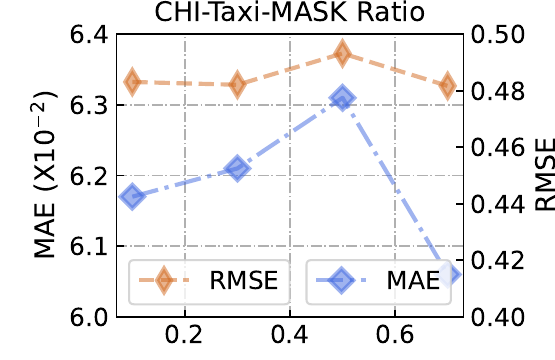}
  \end{minipage}\hspace{-3.0mm}
  &
  
\begin{minipage}{0.24\textwidth}
	\includegraphics[width=\textwidth]{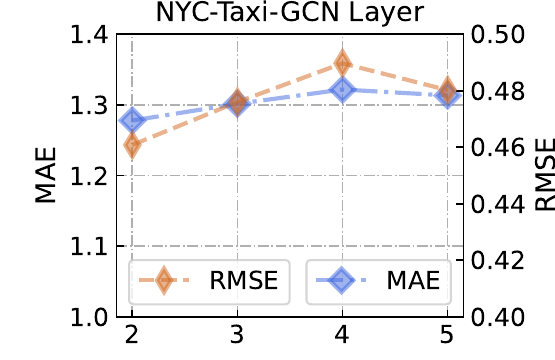}
  \end{minipage}\hspace{-3.mm}
  &
  \begin{minipage}{0.24\textwidth}
	\includegraphics[width=\textwidth]{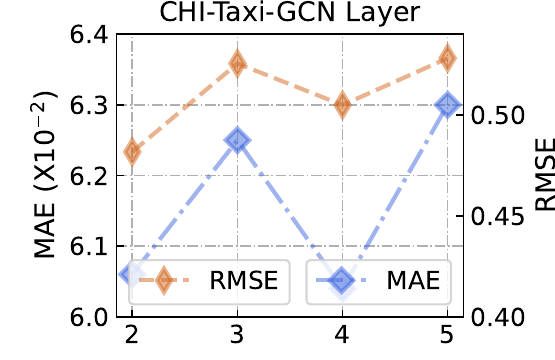}
  \end{minipage}\hspace{-3.0mm}
\end{tabular}
\vspace*{-0.15in}
\caption{Hyperparameter study on traffic prediction}
\vspace*{-0.15in}
\label{fig:hyper_traffic1}
\end{figure*}

\begin{figure}[t]
\vspace*{-4mm}
\centering
\begin{tabular}{c c }
\\\hspace{-4.0mm}
\begin{minipage}{0.23\textwidth}
	\includegraphics[width=\textwidth]{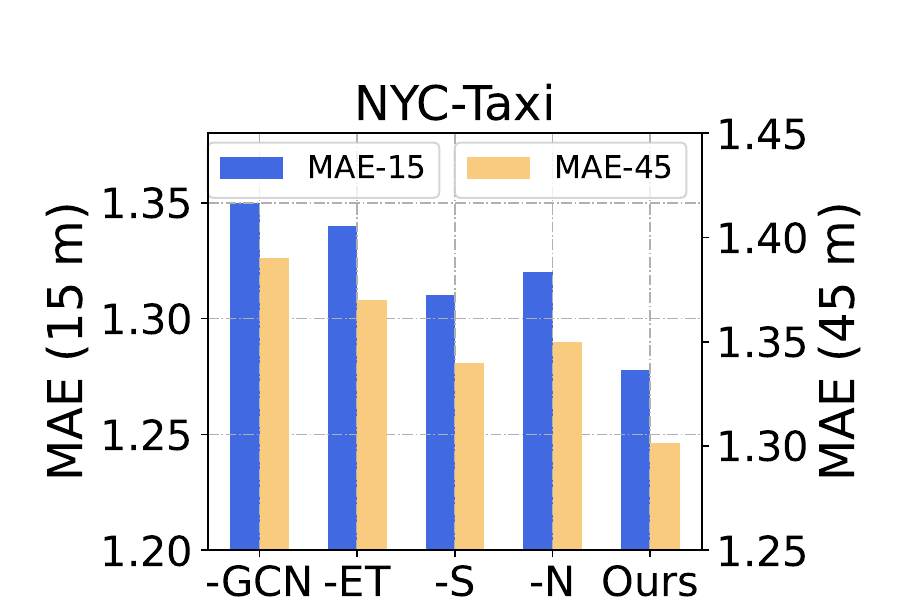}
  \end{minipage}\hspace{-3.mm}
  &
  \begin{minipage}{0.23\textwidth}
	\includegraphics[width=\textwidth]{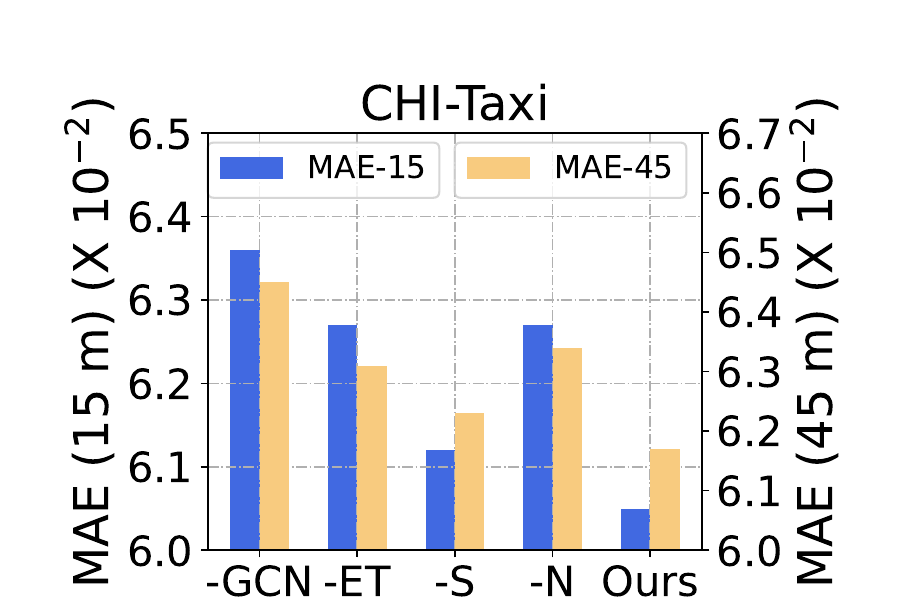}
  \end{minipage}\hspace{-3.mm}
\end{tabular}
\vspace*{-2mm}
\caption{Ablation study of \model\ on traffic prediction}
\vspace*{-3mm}
\label{fig:ablation_traffic}
\end{figure}

\begin{figure}[t]
\centering
\begin{tabular}{c c}
\centering
\hspace{-3.0mm}
  \begin{minipage}{0.23\textwidth}
	\includegraphics[width=\textwidth]{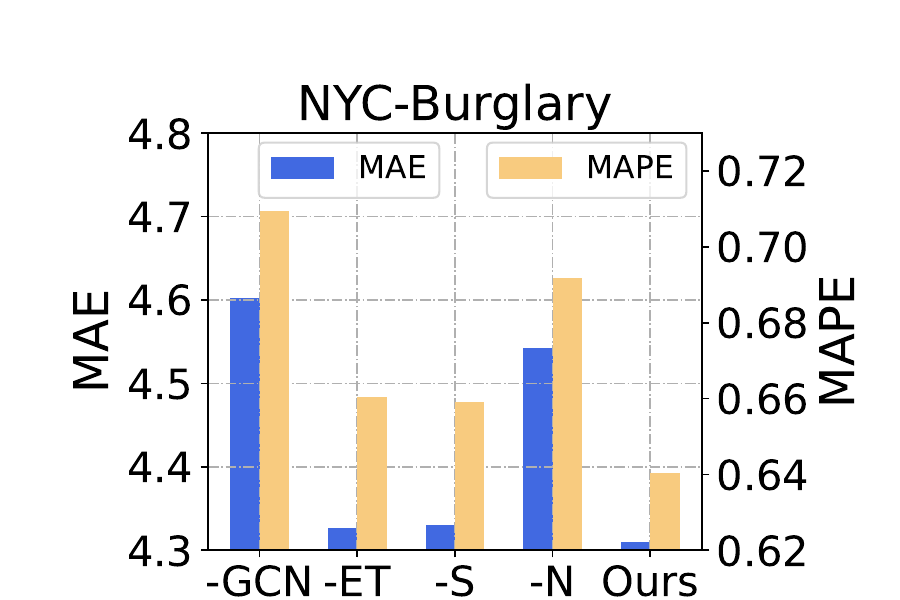}
  \end{minipage}\hspace{-3.5mm}
  &
  \begin{minipage}{0.23\textwidth}
	\includegraphics[width=\textwidth]{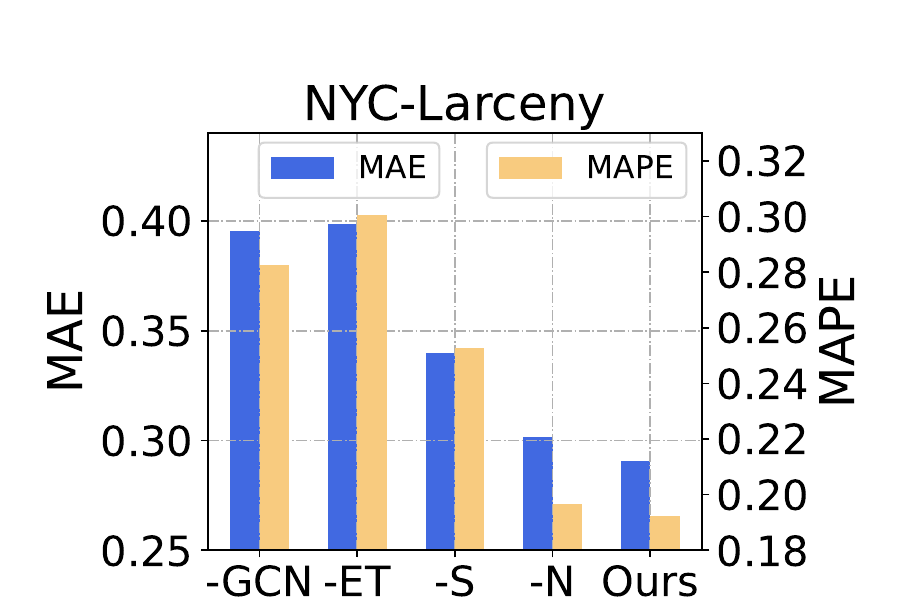}
  \end{minipage}\hspace{-3.5mm}
  \\\hspace{-3.0mm}
  \begin{minipage}{0.23\textwidth}
	\includegraphics[width=\textwidth]{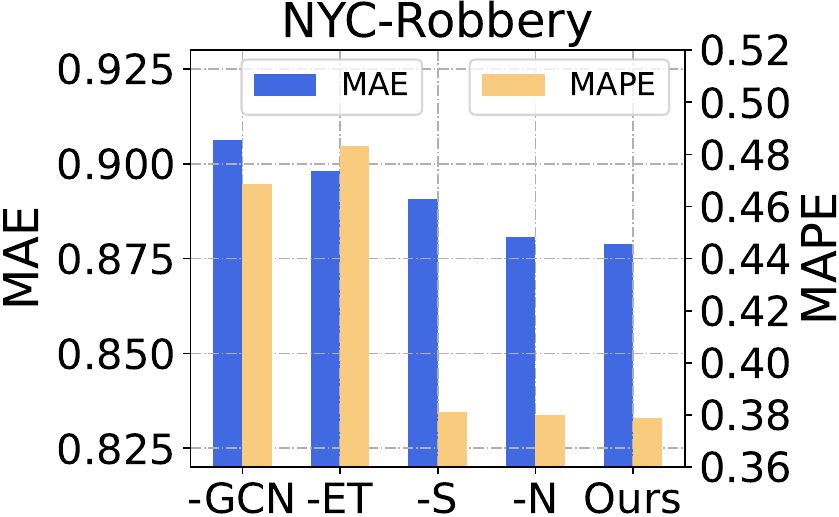}
  \end{minipage}\hspace{-3.0mm}
  &
  \begin{minipage}{0.23\textwidth}
	\includegraphics[width=\textwidth]{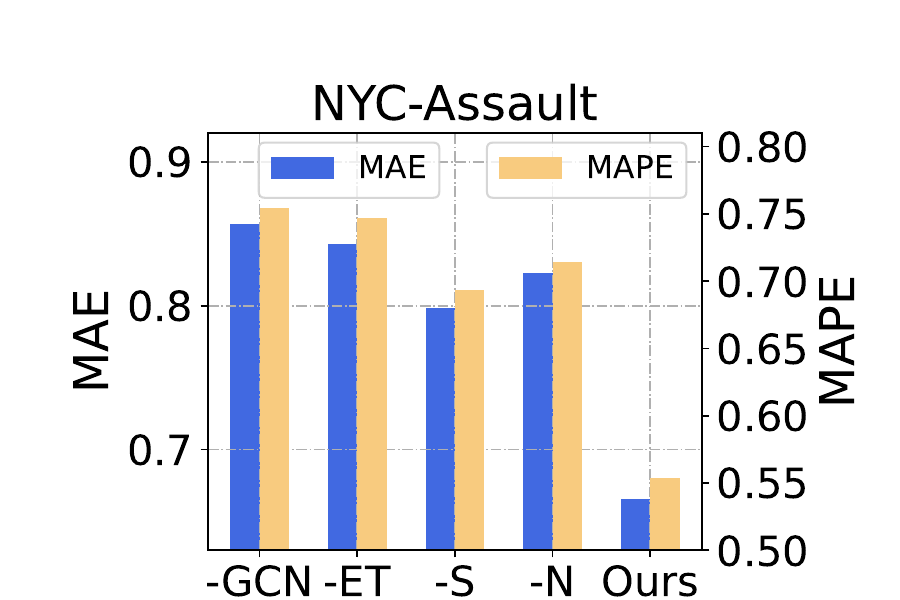}
  \end{minipage}\hspace{-3.5mm}
\end{tabular}
\vspace{-0.1in}
    \caption{Ablation study of \model\ on crime prediction}
    \vspace{-0.2in}
    \label{fig:ablation_crime}
\end{figure}

\subsection{Ablation Study (RQ2)}

To evaluate the effectiveness of each component of our framework \model, we perform an ablation study and present results \wrt~traffic prediction and crime prediction in Figure~\ref{fig:ablation_traffic} and Figure~\ref{fig:ablation_crime}.
Particularly, we conduct an ablation study several variants including ``RP GCN'' by replacing GCN with MLP, ``w/o Structure Mask'', ``w/o Node Mask'', and ``w/o Edge Type Mask'' by removing masking links or masking nodes. 

\noindent \textbf{Analysis on Crime prediction}. The figure demonstrates the impact of different components on region representation for crime prediction. ``RP GCN'' replacing GCN with a 2-layer MLP, contributes the most to performance improvement compared to ``w/o Structure Mask'', ``w/o Edge Type Mask'',  and ``w/o Node Mask'' GCN captures spatial-temporal graph structure semantics, unlike MLP. ``w/o Node Mask'' has a larger impact than ``w/o Structure Mask'' as masking nodes directly affects region representations. Masking different link types positively affects region representations, emphasizing the significance of masking links in the heterogeneous graph. ``w/o Edge Type Mask'' which doesn't mask edges, shows a notable performance decline, highlighting the effectiveness of edge type masking in our \model\ model. Overall, integrating GCN, node masking, and edge-type masking enhances the performance of our \model\ framework.

\noindent \textbf{Analysis on Traffic prediction}. Based on Figure~\ref{fig:ablation_traffic}, GCN has the greatest impact on improving region representations and traffic prediction, especially for longer time horizons. Masking edge types significantly improves traffic flow prediction accuracy. Masking nodes and graph structure also contribute positively by refining region representations and capturing spatial relationships. Incorporating POI information enhances region representations and enables the encoding of different region relations. Overall, the combination of GCN, edge type masking, node masking, and POI integration contributes to the success of our \model\ method in learning region representations and achieving accurate traffic predictions.

\subsection{More Effectiveness Evaluation Results (RQ1)}
Results for the data on home prices in Chicago and New York City are shown in Table~\ref{tab:over_results} and more analysis details are below. A Lasso regression approach uses the input embeddings from the learned region representations created using various methods. 
\begin{itemize}[leftmargin=*]
\item \model\ consistently performs at the highest level. On the other hand, MVURE and MGFN perform better when using data from the housing markets of New York and Chicago, respectively. This is because they mine human patterns, which connect comparable or nearby places. We also notice that the POI technique performs well because it creates connections between POIs of regions and core shopping regions, which include POIs that are distinct from those in remote towns. 
\item Methods taking advantage of region representations outperform those using network embeddings in terms of performance. The basis for this investigation is that region-wise dependencies from both the spatial and temporal dimensions cannot be effectively encoded by conventional graph-based approaches (such as GCN).
\item Table~\ref{table:traffic_add} shows results of improvement of existing widely used traffic prediction methods (\eg, GMAN, STGCN) based on several region representation baselines and our method \model. The superior performance further validates the effectiveness of generative self-supervised learning with a masking mechanism for denoising data. In contrast, methods like MVURE and MGFN require significantly more data to uncover human patterns.
\end{itemize}

\begin{table}
\centering
\scriptsize
\caption{Overall Performance Comparison in Traffic Prediction Improvement}
\label{table:traffic_add}
\setlength{\tabcolsep}{2.8mm}{
\begin{tabular}{|c|cc|c|cc|}
\hline
\multirow{2}{*}{Methods} & \multicolumn{2}{c|}{NYC-Taxi}   & \multirow{2}{*}{Methods} & \multicolumn{2}{c|}{NYC-Taxi}   \\ \cline{2-3} \cline{5-6} 
                         & \multicolumn{1}{c|}{MAE} & RMSE &                          & \multicolumn{1}{c|}{MAE} & RMSE \\ \hline
GMAN                     & \multicolumn{1}{c|}{2.0812}    &3.0215      & STGCN                    & \multicolumn{1}{c|}{1.4093}    &4.1766      \\ \hline
w/MVURE                  & \multicolumn{1}{c|}{1.8795}    &2.8337      & w/MVURE                  & \multicolumn{1}{c|}{1.3007}    &3.6715      \\ \hline
w/MGFN                   & \multicolumn{1}{c|}{1.9261}    &2.8337      & w/MGFN                   & \multicolumn{1}{c|}{1.3266}    &3.7514      \\ \hline
w/\model                       & \multicolumn{1}{c|}{1.8403}    &2.7032       & w/\model                       & \multicolumn{1}{c|}{1.2415}    &3.6375      \\ \hline
\end{tabular}}
\end{table}

\vspace{-0.1in}
\subsection{Model Robustness Study (RQ3)}
We also conduct experiments to test how well our framework \model\ stands up to data sparsity. To accomplish this, we assess the forecast accuracy of regions with various densities independently. Here, the ratio of non-zero elements (crime occurs) in the region-specific crime occurrence sequence $\mathbf{Z}_r$ is used to estimate the density degree of each region. We divide sparse areas with a crime density degree of less than 0.5 into two groups, (0.0, 0.25] and (0.25, 0.5] specifically. Figure~\ref{fig:robustness1} presents the evaluation findings. We see that our \model\ constantly performs better than alternative approaches. Thus, this experiment once more indicates how masking mechanism on nodes and the structure for self-supervised learning into the spatial-temporal region representation learning provides very accurate and reliable crime prediction in all scenarios with varying crime density degrees. To mine human behaviors or movement patterns, existing region representation algorithms (such MVURE and MGFN) require a huge quantity of data, which results in lower performance on the sparse data. Another finding acknowledges that typical graph learning techniques now in use (such as GNN-based approaches including GCN) are unable to efficiently build high-quality representations of regions with sparse crime data via the spatial-temporal graph.

The sencond line of Figure~\ref{fig:robustness1} present more details about the evaluation findings of the model robustness. We see that \model\ constantly performs better than alternative approaches. Thus, this experiment once more indicates how masking mechanism on nodes and the structure for self-supervised learning into the spatial-temporal region representation learning provides very accurate and reliable crime prediction in all scenarios with varying crime density degrees. To mine human behaviors or movement patterns, existing region representation algorithms (MVURE and MGFN) require a huge quantity of data, which results in lower performance on the sparse data. Another finding acknowledges that typical graph learning techniques now in use (\textit{i.e.,} GNN-based approaches like CGN) are unable to efficiently build high-quality representations of regions with sparse crime data via the spatial-temporal graph.

\subsection{Hyperparameter Study (RQ4)}
We conduct the hyperparameter study on the parameters including the mask ratio of nodes and links with the range $[0.1,0.3,0.5,0.7]$, and the number of CGN layer from the range $[2,3,4,5]$ on crime prediction task and traffic predition ~\wrt~New York and Chicago in terms of MAE, MAPE and RMSE. We provide results in Figure~\ref{fig:hyper_traffic1}. From figures, we can have the following observations:
\textbf{(1)} For the impact of mask ratio of nodes and structure, we find that the model \model\ can achieve the best performance on crime prediction and traffic prediction on all datasets when the mask ratio of nodes and links is set as 0.7. And the performance decreases with the decrease of mask ratio. This verifies that masking nodes and links do not always bring better model representation ability. Smaller mask ratios (\eg, $[0.1, 0.3, 0.5]$) may not denoise nosiy nodes and links completely, which leads to bad performance. 
\textbf{(2)} For the effect of the number of GCN layer, we find that our framework \model\ achieves the best performance when the number of GCN layer is set as 2. We notice that the performance decreases with the increasing number of GCN layer. This may be caused by over-smoothing issue on region representation with the increasing number of the GCN layer. 
\textbf{(3)} We also investigate the consuming time of our model \model\ with varied hyperparameters. We observe that time cost increases with increasing mask ratio of nodes and links, since our model \model\ requires to reconstruct more nodes' representation and new links when masking more nodes and links. Similarity, time consuming is larger when the more layers of GCN due to that the message passing between GCN layers need more time.

\subsection{Model Efficiency Study (RQ5)}
We investigate the efficiency of our method \model\ and existing region representation learning methods ~\wrt~training time. We provide results in Table~\ref{table:efficiency}. All of these methods are adopted in python 3.8, pytorch 1.7.0 with cuda 11.3 and tensorflow 1.15.3 (GPU version). The experiments are conducted in the same server. From the table, By comparing the model efficiency of \model\ to the baselines (\eg, HDGE, MVURE and MGFN), we can see that it is competitive, indicating that it has the capacity to handle massive amounts of spatial-temporal data. The performance of the region representation is greatly improved by the generative self-supervised learning paradigms while incurring little additional expense.

\begin{table}
\vspace{-0.1in}
\center
\setlength{\tabcolsep}{3.0pt}
\footnotesize
\caption{Computational cost (seconds) of our \model\ and SOTA spatial-temporal region representation methods. MAE and MAPE are estimated for NYC crime prediction results.}
\vspace{-0.12in}
\label{table:efficiency}
\resizebox{\linewidth}{!}{
\begin{tabular}{|c|c|c|c|c|c|c|c|c}
\hline
Models & HDGE & ZE-Mob & MV-PN & CGAL & MVURE & MGFN  &Ours \\ \hline
Training        &303.7     &82.7       &33.4      &4144.8       &240.7       &852.3      &124.5    \\\hline
MAE        &4.2021     &4.3560        &4.2342       &4.1782       &3.7683       &3.5971        &2.1060    \\\hline
MAPE        &0.7821     &0.8012        &0.7791       &0.7506       &0.7318       &0.7065        &0.5203    \\\hline
\end{tabular}
}
\vspace{-0.25in}
\end{table}

\section{Related Work}
\label{sec:relate}
\noindent\textbf{Self-supervised Learning for Spatial-Temporal Graphs}. In the domain of self-supervised learning (SSL) for spatial-temporal graphs, a line of research has emerged, focusing on graph-based SSL methods specifically designed for capturing the spatial and temporal dependencies within the graph structure. Several recent studies have contributed to this area, including ~\cite{liu2023self,ji2023spatio,zhang2023automated,zhang2023spatial}, and ~\cite{ji2022self}. Liu et al.~\cite{liu2023self} propose an innovative approach that adopts the self-supervised learning paradigm to leverage supplementary information that is typically inaccessible due to privacy concerns. 
Ji et al.~\cite{ji2023spatio} introduce self-supervised learning paradigms in a spatial-temporal traffic prediction framework. The proposed method enhances the traffic pattern representations to capture both spatial and temporal heterogeneity within the spatial-temporal graph, leading to improved traffic prediction performance. Overall, these studies demonstrate the effectiveness of methods of self-supervised learning in capturing spatial and temporal dependencies within spatial-temporal graphs. By leveraging additional information or incorporating self-supervised learning paradigms, these approaches enhance representation learning and improve the performance of downstream tasks in various domains, including POI recommendation, traffic prediction, and region representations on the spatial-temporal graph for downstream tasks.

\noindent\textbf{Graph Mask Autoencoder}. Graph autoencoder methods have emerged as a category of generative self-supervised learning approaches that aim to construct input graphs directly, eliminating the need for high-quality data augmentation typically required in contrastive self-supervised learning. An earlier study by Kipf and Welling~\cite{kipf2016variational} introduces Graph Autoencoder (GAE) and Variational Graph Autoencoder (VGAE), which utilize Graph Convolutional Networks (GCN) as the encoder and a dot-product decoder. These models focus on reconstructing the topological structure and node content of the graph from a compressed latent representation. ARVGA (Adversarially Regularized Variational Graph Autoencoder)~\cite{pan2018adversarially} proposes a novel adversarial graph embedding framework. It trains a decoder to reconstruct both the topological structure and node content of the graph from a compact representation generated by the framework. This adversarial training enhances the robustness and generalization capabilities of the model. MGAE (Multi-modal Graph Autoencoder)\cite{wang2017mgae} and GALA (Graph Autoencoder with a Ladder-like Architecture)\cite{park2019symmetric} are methods that combine structure and node reconstruction. They aim to learn a comprehensive representation of the graph by simultaneously reconstructing the graph structure and node attributes. In a recent study by Hou et al.~\cite{hou2022graphmae}, a graph autoencoder approach called GraphMAE is proposed, which focuses solely on node reconstruction for general graphs. By reconstructing nodes, GraphMAE achieves promising results in node classification tasks.

\noindent\textbf{Region Representation Learning}. Research on the region representation learning problem includes the following studies~\cite{wang2017region,yao2018representing,zhang2021multi,zhang2019unifying,fu2019efficient,wu2022multi_graph,zhang2023automated}. In particular, Fu et al.~\cite{fu2019efficient} suggest to incorporate both intra-region information (such as POI distance inside a region) and inter-region information (such as the similarity of POI distributions between two regions) to increase the quality of region representations. The concept of putting forward in~\cite{fu2019efficient} is expanded upon by Zhang et al.~\citep{zhang2019unifying} by using a collective adversarial training.

\noindent In a recent study conducted by Zhang et al.~\cite{zhang2021multi}, a multi-view joint learning model is developed to learn region representations. To capture region correlations from different perspectives, such as human mobility and region properties, the study incorporates a graph attention technique for each view. However, this approach focuses solely on modeling region correlations and does not consider the inclusion of point-of-interest (POI) data, which is essential for capturing region functionality. Similarly, Wu et al.~\cite{wu2022multi_graph} propose a method for learning area representations by extracting traffic patterns. However, their approach also excludes POI data and relies solely on mobility data. The efficacy of these techniques heavily relies on the generation of high-quality region graphs. Furthermore, they may struggle to learn accurate region representations when faced with noisy and skewed spatio-temporal data. In contrast, a recent study by Zhang et al.~\cite{zhang2023automated} proposes the adoption of a contrastive self-supervised learning paradigm on a multi-view heterogeneous spatio-temporal graph. This approach utilizes a self-contrastive learning method to obtain high-quality region vectors. By contrast, our paper employs a generative self-supervised learning method to model region correlations by masking links and nodes on the heterogeneous spatio-temporal graph.

\section{Conclusion}
\label{sec:conclusoin}
We introduce a novel generative spatial-temporal graph self-supervised learning paradigm for region representation. Our approach incorporates mask mechanisms to address limitations in existing area representation methods. Through extensive experiments on real-world datasets and three spatial-temporal mining tasks, we demonstrate the effectiveness of our \model\ in learning area graph representations from multi-view spatial-temporal data. Our contributions include proposing a self-supervised learning paradigm using generative models and masks, addressing challenges in area representation methods, and achieving superior performance compared to state-of-the-art approaches in various spatial-temporal mining tasks, such as crime prediction.

\section*{Acknowledgement}
\label{sec:ack}
The Australian Research Council partially supports this work under the streams of Future Fellowship (Grant No. FT210100624),  the Discovery Project (Grant No. DP240101108), and the Linkage Projects (Grant No. LP230200892 and LP240200546).
This project is also partially supported by Shenzhen-Hong Kong-Macao Science and Technology Plan Project (Category C Project: SGDX202108 23103537030) and Theme-based Research Scheme T35-710/20-R. We appreciate help of Dr. Huang and Dr. Xia.

\bibliographystyle{ACM-Reference-Format}
\bibliography{sample-base}

\end{document}